\documentclass[conference]{IEEEtran}
\IEEEoverridecommandlockouts
\usepackage[noadjust]{cite}
\usepackage{amsmath,amssymb,amsfonts}
\usepackage{algorithmic}
\usepackage{graphicx}
\usepackage{textcomp}
\usepackage{xcolor}
\usepackage{times}
\usepackage{multirow}
\usepackage{multicol}
\def\BibTeX{{\rm B\kern-.05em{\sc i\kern-.025em b}\kern-.08em
    T\kern-.1667em\lower.7ex\hbox{E}\kern-.125emX}}

\makeatletter
\renewcommand\IEEEkeywordsname{Keywords}
\makeatother
\usepackage{xpatch}
\xpatchcmd\IEEEkeywords{---}{-}{}{}

\makeatletter
\renewcommand{\fnum@figure}{Figure~\thefigure}
\makeatother

\begin{document}

\title{\bfseries\Large Intermediate-Task Transfer Learning: Leveraging Sarcasm Detection for Stance Detection

}

\author{\IEEEauthorblockN{Gibson Nkhata, Susan Gauch}
\IEEEauthorblockA{\textit{Department of Electrical Engineering \& Computer Science} \\
\textit{University of Arkansas}\\
Fayetteville, AR 72701, USA \\
Emails: gnkhata@uark.edu, sgauch@uark.edu}
}

\maketitle

\begin{abstract}
Stance Detection (SD) in the context of social media has emerged as a prominent area of interest with implications for social, business, and political applications, thereby garnering escalating research attention within the realm of Natural Language Processing (NLP). The inherent subtlety, nuance, and complexity of texts procured from online platforms via crowd-sourcing pose challenges for SD algorithms in accurately discerning the author's stance. Particularly, the inclusion of sarcastic and figurative language drastically impacts the performance of SD models. This paper addresses this challenge by employing sarcasm detection intermediate-task transfer learning tailored for SD. The proposed methodology involves the fine-tuning of BERT and RoBERTa and the sequential concatenation of convolutional, bidirectional LSTM, and dense layers. Rigorous experiments are conducted on publicly available benchmark datasets to evaluate our transfer-learning framework. The performance of the approach is assessed against various State-Of-The-Art (SOTA) baselines for SD, providing empirical evidence of its effectiveness. Notably, our model outperforms the best SOTA models, achieving average F1-score gaps of 0.038 and 0.053 on the SemEval 2016 Task 6A Dataset (SemEval) and Multi-Perspective Consumer Health Query  Data (MPCHI), respectively, even prior to sarcasm-detection pre-training. The integration of sarcasm knowledge into the model proves instrumental in mitigating misclassifications of sarcastic textual elements in SD. Our model accurately predicts 85\% of texts that were previously misclassified by the model without sarcasm-detection pre-training, thereby amplifying the average F1-score of the model. Furthermore, our experiments revealed that the success of the transfer-learning framework is contingent upon the correlation of lexical attributes between the intermediate task (sarcasm detection) and the target task (SD). This study represents the first exploration of sarcasm detection as an intermediate transfer-learning task in the context of SD and simultaneously exploits the concatenation of BERT or RoBERTa with other deep-learning techniques, establishing the proposed approach as a foundational baseline for future research endeavors in this domain.
\end{abstract}


\section{Introduction}
Social media platforms, increasingly popular, enable individuals to freely express opinions and connect globally for real-time updates on diverse topics~\cite{savini2022intermediate,ghosh2019stance,aldayel2021stance}. Discourse on emerging subjects yields substantial data valuable for Natural Language Processing (NLP) tasks, notably Stance Detection (SD). SD is the automated identification of an individual's stance on a specific topic based solely on their utterance or authored material~\cite{ghosh2019stance}\cite{augenstein2016stance,kuccuk2020stance,kuccuk2022tutorial}. Stance labels categorize expressions into \textit{InFavor}, \textit{Against}, or \textit{None}. This phenomenon, particularly on social media, is a burgeoning focus in social, business, and political applications~\cite{aldayel2021stance}\cite{biber1988adverbial}.

Previous SD research has been evaluated using the publicly available datasets crowd-sourced from online platforms~\cite{ghosh2019stance}\cite{aldayel2021stance}\cite{kuccuk2020stance}\cite{kuccuk2021stance}. However, texts procured from online platforms are often characterized by subtlety, nuance, and complexity, featuring inherent sarcastic and figurative language. This complexity poses challenges for SD algorithms in accurately discerning the author's stance~\cite{ghosh2019stance}. Additionally, targets are not consistently mentioned in text~\cite{augenstein2016stance}, and stances are not explicitly transparent. Consequently, inferring the author's stance becomes further complicated, often necessitating implicit inference through a combination of interaction, historical context, and social linguistic attributes, such as sarcasm or irony. 

Prior work has explored intermediate-task transfer learning, involving fine-tuning a model on a secondary task before its application to the primary task to address the aforementioned challenge~\cite{savini2022intermediate}\cite{phang2018sentence,li2019multi,sap2019socialiqa,pruksachatkun2020intermediate,hardalov2022few}. Specifically,~\cite{li2019multi} and~\cite{hardalov2022few} utilized sentiment classification to enhance their models for SD. In a similar vein,~\cite{savini2022intermediate} incorporated emotion and sentiment classification prior to sarcasm detection. The study by~\cite{savini2022intermediate} suggested that pre-training a model with sentiment analysis before sarcasm detection enhances overall performance, attributing this improvement to the correlation between sarcasm and an implied negative sentiment. This finding aligns with one of our experimental observations in Section \ref{Exp}, wherein most sarcastic sentences with an ``Against" stance were initially misclassified as ``InFavor" before the integration of sarcasm pre-training into our model. Nonetheless, sarcasm language in the target tasks has detrimentally affected performance, and previous research has not explored the sarcasm phenomenon for enhancing SD models. In this study, our focus is to experiment with and employ sarcasm detection as an intermediate task tailored to improve SD performance.

 Sarcasm detection involves inferring intention or secondary meanings from an utterance, discounting literal meaning~\cite{ghosh2016fracking}. It employs positive words and emotions to convey negative or undesirable figurative attributes, serving as a mechanism to express opinions using seemingly conflicting language ~\cite{sarsam2020sarcasm,jamil2021detecting,8949523}. Sarcasm can alter the stance of a text from \textit{Against} to \textit{InFavor} and vice versa~\cite{jamil2021detecting}\cite{liebrecht2013perfect}. Thus, we propose infusing sarcasm knowledge into the model before SD fine-tuning to enhance performance. 

This work employs a model framework consisting of BERT~\cite{devlin2018bert} or RoBERTa~\cite{liu2019roberta}, two convolutional layers (Conv), a Bidirectional LSTM layer (BiLSTM), and a dense layer. Experimental results affirm the efficacy of our approach, demonstrated by improved F1-scores upon the inclusion of sarcasm detection in the model framework. Furthermore, the significance of this approach is emphasized by presenting a sample of sarcastic texts from datasets during a failure analysis of the original SD model results, prior to the incorporation of sarcasm intermediate-task pre-training. Exploring three publicly available sarcasm datasets, we find that different sarcasm detection tasks impact SD performance variably, depending on linguistic and quantitative attributes. Our work makes the following key contributions:

\begin{itemize}
\item
\textit{Transfer-Learning Framework:} Introducing a novel transfer-learning framework incorporating sarcasm detection as an intermediate task before fine-tuning on SD, utilizing an integrated deep learning model.

\item
\textit{Performance Superiority:} Demonstrating superior performance against State-Of-The-Art (SOTA) SD baselines, even without sarcasm detection pre-training, indicated by higher F1-scores..

\item
\textit{Correlation Analysis:} Establishing and illustrating the correlation between sarcasm detection and SD, exemplified through a failure analysis, thereby emphasizing the improvement of SD through sarcasm detection.

\item
\textit{Impact Assessment:} Measuring the impact of various sarcasm detection models on target tasks based on the correlation between linguistic and quantitative attributes in the datasets of the two tasks.

\item
\textit{Ablation Study:} Conducting an ablation study to assess the contribution of each module to the overall model framework. The study also reveals a significant drop in performance without sarcasm knowledge, underscoring the importance of our proposed approach.
\end{itemize}

The remainder of this paper unfolds as follows: Section~\ref{Rel} reviews related work, Section~\ref{Meth} outlines our proposed approach, and Section~\ref{Exp} delves into comprehensive experiments, covering datasets, results, and subsequent discussions. The conclusion and recommendations for further study are provided in Section~\ref{Con}. The final section critically examines the limitations inherent in our study.

\section{Related work} \label{Rel}
This section conducts a literature review on SD and intermediate-task transfer learning.

\subsection{Stance Detection (SD)}
The literature on SD has traditionally explored two primary perspectives: Target-Specific SD (TSSD), focusing on individual targets~\cite{ghosh2019stance}\cite{aldayel2021stance}\cite{zhang2022would}\cite{liang2022zero}, and Multi-Target SD (MTSD), concurrently inferring stances towards multiple related subjects~\cite{liang2022zero,sobhani2017dataset,liu2018two,sobhani2019exploring}. Early SD approaches utilized rule-based methods~\cite{zhang2022would}\cite{walker2012stance}, followed by classical machine learning techniques~\cite{kuccuk2018stance}\cite{segura2018labda}. Later, the emergence of deep learning models led to neural networks supplanting classical approaches~\cite{augenstein2016stance}\cite{hardalov2022few}\cite{siddiqua2019tweet}\cite{ng2022my}. For instance, a neural ensemble model incorporating BiLSTM, attention mechanism, and multi-kernel convolution was presented in~\cite{siddiqua2019tweet}, evaluated on both TSSD and MTSD. While our work shares similarities in model framework, it distinctively employs BERT or RoBERTa and introduces an intermediate-task transfer learning technique, deviating from ensemble approaches and multi-kernel usage. 

Recent efforts have explored the use of pre-trained language models for SD. While~\cite{ghosh2019stance} conducted a comparative study, fine-tuning pre-trained BERT against classical SD approaches,~\cite{liang2022zero} employed BERT as an embedding layer to encode textual features in a zero-shot deep learning setting, yielding promising results; however, both studies observed challenges in accurately classifying sarcastic examples. On the other hand,~\cite{zhang2022would} experimented with ChatGPT, prompting the model directly with test cases to discern their stances.

\subsection{Intermediate-Task Transfer Learning}
Recent studies have also embraced intermediate-task transfer learning to transfer knowledge from a data-rich auxiliary task to a primary task~\cite{pruksachatkun2020intermediate}. This technique has shown significant success in various NLP tasks. For instance,~\cite{phang2018sentence} employed supervised pre-training with four-example intermediate training tasks to enhance performance on the primary task evaluated using the GLUE benchmark suite~\cite{wang2018glue}. Furthermore, ~\cite{hardalov2022few} introduced few-shot learning, utilizing sentiment-based annotation to improve cross-lingual SD performance. Additionally,~\cite{savini2022intermediate} employed transfer learning by separately fine-tuning pre-trained BERT on emotion and sentiment classification before fine-tuning the model on the primary task of sarcasm detection, leveraging the correlation between sarcasm and negative sentiment polarity.

To our knowledge, prior work has not explored sarcasm detection pre-training for SD, nor has it investigated the concatenation of BERT or RoBERTa with other deep-learning techniques for SD. In this paper, we propose leveraging sarcasm detection for TSSD  within a model framework comprising BERT, Conv, BiLSTM, and a dense layer.

\section{Methodology} \label{Meth}
This section outlines our approach, encompassing the intermediate-task transfer learning and the underlying model architecture.

\subsection{Intermediate-Task Transfer Learning}

Our model adopts a single intermediate-task training, which consists of two phases: pre-training on an intermediate task and fine-tuning on a target task.

\subsubsection{Target Task}
The focal task in this study is SD, where the objective is to predict the stance expressed in a given text, such as a tweet, towards a specified target, like  `\textit{Feminist Movement}'. A tweet, denoted as $T$, is represented as a word sequence ($w_{1},w_{2},w_{3},...w_{L}$), with \textit{L} denoting the sequence length. Stance labels are categorized as  \textit{InFavor} (supporting the target/topic/claim), \textit{Against} (opposing the topic), or \textit{None} (indicating neutrality towards the target).

\subsubsection{Intermediate Task}
The intermediate task in this study is sarcasm detection. As prior research has not employed sarcasm as an intermediate task, we investigate the following three sarcasm-detection tasks to gain insights into the crucial linguistic attributes for a model to learn from the intermediate task, aiming to enhance SD performance. 

\textit{Sarcasm V2 Corpus (SaV2C).}
The SaV2C dataset, introduced by~\cite{oraby2017creating}, presents a diverse corpus of sarcasm, utilizing syntactical cues and crowd-sourced from the Internet Argument Corpus (IAC 2.0). It comprises 4,692 lines extracted from Quote and Response sentences in political debate dialogues. Our exploration focuses on the General Sarcasm category within the dataset, containing 3,260 instances each of sarcastic or non-sarcastic comments.

\textit{The Self-Annotated Reddit Corpus (SARC).}
The SARC dataset~\cite{khodak2018large} is derived from Reddit. In contrast to the other datasets, sarcasm annotations in SARC are directly provided by the authors, ensuring reliable and trustworthy data. Due to accessibility issues with the original website, we obtained the Main Balanced first version of the dataset directly from the author of~\cite{savini2022intermediate}. This version comprises 1,010,826 training samples, evenly distributed between sarcastic and non-sarcastic instances.

\begin{figure}[h]
\center
\includegraphics[width=0.48\textwidth]{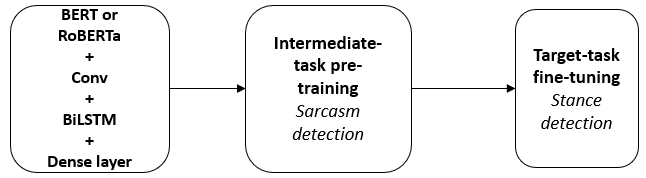}
\caption{Intermediate-task transfer learning pipeline.} 
\label{fig:exp_ppl}
\end{figure}

\textit{SARCTwitter (ST)}.
The ST dataset~\cite{mishra2016predicting} is designed to predict readers' sarcasm understandability using features, including eye movement. In our study, we utilized the dataset variant employed by~\cite{majumder2019sentiment}, excluding the eye movement feature. Crowd-sourced from Twitter (X), ST is manually annotated by seven readers and contains 350 sarcastic and 644 non-sarcastic tweets. Our intermediate-task transfer learning pipeline is depicted in Figure~\ref{fig:exp_ppl}.

\subsection{Underlying Model Architecture}
The entire model framework primarily comprises an input layer, an embedding layer, and deep neural networks. 

\subsubsection{Input Layer}
This layer takes a text \textit{S} encoding the stance information and comprising \textit{n} words.  \textit{S} is transformed into a vector of words and passed to the embedding layer.

\subsubsection{The Embedding Layer}

We employ BERT~\cite{devlin2018bert} and RoBERTa~\cite{liu2019roberta} for textual input encoding into hidden state \textit{H} in our experimentation. Noteworthy achievements of these language models in the literature ~\cite{savini2022intermediate}\cite{ghosh2019stance}\cite{phang2018sentence}\cite{pruksachatkun2020intermediate}\cite{ng2022my}\cite{gibthesis} motivate their exploration to identify the most suitable model for alignment with our research objectives.

\subsubsection{Deep Neural Networks}
\begin{figure*}[ht]
\center

\includegraphics[width=0.95\textwidth]{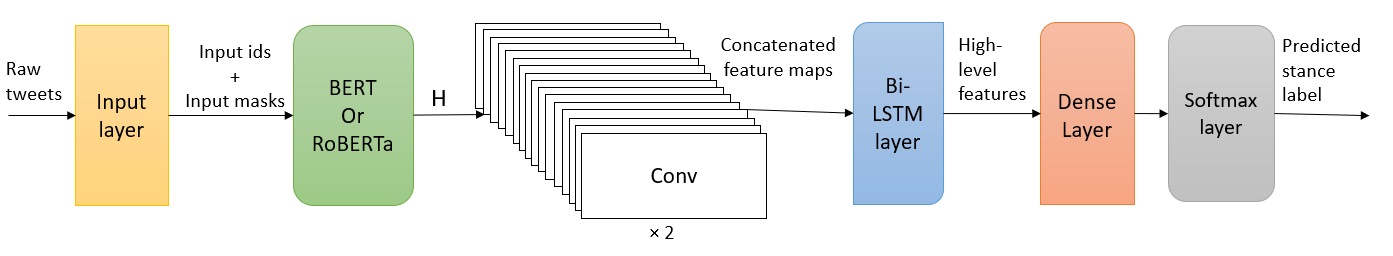}
\caption{Proposed model framework.} 
\label{fig:model}
\end{figure*}

This module utilizes Conv, a BiLSTM layer, and a dense layer, positioned atop the embedding layer. The purpose of incorporating convolution is to discern specific sequential word patterns within a sentence, generating a composite feature map from \textit{H}. This feature map facilitates the BiLSTM layer in acquiring nuanced higher-level stance representations, which are subsequently mapped into a more differentiable space by the dense layer. Figure \ref{fig:model} depicts the overall model framework.

\section{Experiments} \label{Exp}
This section delineates the datasets employed, details the data pre-processing procedures, outlines baseline models, presents experimental results, and engages in a subsequent discussion.

\subsection{Datasets}
For evaluation purposes, we employed two publicly available SD datasets: 1) the well-established SemEval 2016 Task 6A Dataset (SemEval); and 2) the Multi-Perspective Consumer Health Query Data (MPCHI). 

\subsubsection{SemEval}
The SemEval~\cite{MohammadSK17} task encompasses tweets manually annotated for stance towards a specified target, a target of opinion, and sentiment. Our experiments exclusively utilize tweets and their corresponding stance annotations. The dataset comprises tweet data associated with five distinct targets: Atheism (AT), Climate Change (CC), Feminist Movement (FM), Hillary Clinton (HC), and Legalization of Abortion (LA). 

\subsubsection{MPCHI}
MPCHI~\cite{sen2018stance} serves as a dataset for stance classification to enhance Consumer Health Information (CHI) query search results. Comprising formal texts extracted from top-ranked articles corresponding to queries on a specific web search engine, the dataset includes sentences related to five distinct queries, which are also the targets for stance classification: MMR vaccination can cause autism (MMR), E-cigarettes are safer than normal Cigarettes (EC), women should take HRT post menopause (HRT), Vitamin C prevents common cold (VC), and Sun exposure leads to skin Cancer (SC).

\begin{table*}[ht]
\caption{Statistics of the datasets divided into training and test sets}
\begin{center}
\begin{small}
\begin{tabular}{|c||c||c|c|c||c|c|c|}
\hline
\multirow{2}{*}{\textbf{Dataset}}& \multirow{2}{*}{\textbf{Target}}&\multicolumn{3}{|c|}{\textbf{Training samples}} &\multicolumn{3}{|c|}{\textbf{Test samples}}\\
\cline{3-8} 
\textbf{}& &INFAVOR & AGAINST&NONE  &INFAVOR &AGAINST &NONE  \\
\hline
\multirow{5}{*}{SemEval} &AT &92 &304 &117 &32 &160 &28 \\
\cline{2-8}
&CC &212 &15 &168 &123 &11&35\\
\cline{2-8}
&FM  &210 &328 &126 &58 &183 &44\\
\cline{2-8}
&HC &112 &361 &166 &45 &172 &78\\
\cline{2-8}
&LA &105 &334 &164 &46 &189 &45\\
\hline
\hline
\multirow{5}{*}{MPCHI} &MMR &48 &61 &72 &24 &33 &21\\
\cline{2-8}
&SC &68 &51 &117 &35 &26 &42\\
\cline{2-8}
&EC &60 &118 &111 &33 &47 &44\\
\cline{2-8}
&VC &74 &52 &68 &37 &16 &31\\
\cline{2-8}
&HRT &33 &95 &44 &9 &41 &24\\
\hline
\end{tabular}
\end{small}
\end{center}
\label{tab:data}
\end{table*}
%\end{multicols}

Consistent with~\cite{ghosh2019stance}, the datasets are partitioned into training and test sets following similar proportions. Each text in the datasets is annotated with one of three classes: \textit{InFavor}, \textit{Against}, and \textit{None}. Table~\ref{tab:data} provides statistical details describing the datasets.

\subsection{Data pre-processing}
 We conducted standard data pre-processing steps, including case-folding, stemming, stop-word removal, and deletion of null entries, across all datasets. Text normalization, following the approach by~\cite{han2011lexical}, and hashtag pre-processing, using Wordninja~\cite{wninja}, were also performed. However, for neural network models relying on pre-trained embeddings, stemming and stopword removal were omitted, as stemmed versions of terms might not be present in the pre-trained embeddings. The default tokenizer for the corresponding pre-trained language model was employed to tokenize words in tweets before supplying them to the classifier.

\subsection{Baseline models}
Our model is evaluated against the top-performing results from the SemEval challenge~\cite{mohammad2016semeval}, as reproduced in~\cite{ghosh2019stance} with minor modifications. Additionally, we compare our model's performance with the most recent SOTA methods in SD.

\subsubsection{SemEval models}
We select the Target-Specific Attention Neural Network (TAN-) proposed by~\cite{du2017stance}, and the 1-D sem-CNN introduced by~\cite{kim2014convolutional}. Additionally, we adopt Com-BiLSTM and Com-BERT, implementations provided solely by~\cite{ghosh2019stance}.

\subsubsection{ChatGPT and ZSSD}
The work by~\cite{zhang2022would} investigated ChatGPT for SD by directly probing the generative language model for the stance of a given piece of text, focusing on the SemEval task with specific targets: FM, LA, and HC. On the other hand, the Zero-Short SD (ZSSD) technique~\cite{liang2022zero}, employing contrastive learning, was similarly implemented on SemEval only.

\subsection{Experimental settings}

The inductive approach to transfer learning was applied to the entire model framework, initializing the target task model with parameters learned during sarcasm-detection pre-training. Given the primary focus on enhancing model efficacy for the target task, intermediate tasks were divided into training and validation sets solely for model pre-training on sarcasm detection. In contrast, the target task featured a separate test set for final evaluations and comparisons. As Sav2C and ST are the smallest intermediate-task datasets, five-fold cross-validation was employed on both, while SARC, with its larger size, undergoes an 80/20 train/validation split.

A kernel of size 3, 16 filters, and a ReLu activation function have been employed for the convolutional layer. The BiLSTM layer has been used with a hidden state of 768, matching the hidden state size of the pre-trained language models. The dense layer has employed an output size of 3 and a softmax activation function. All experiments have been conducted on an NVIDIA Quadro RTX 4000 GPU.

Hyperparameter tuning has been  performed through multiple experiments, selecting the best intermediate-task training scheme based on holdout development set results. The optimal per-task model has been then evaluated on the test set. Iterating over datasets with a mini-batch of 16 samples, the Adam optimizer~\cite{kingma2014adam} has been used for parameter learning, employing cross-entropy loss as the cost function. Training runs span 10 to 50 epochs, with early stopping triggered if validation accuracy on holdout data stagnates for five consecutive epochs. The training schedule involves an initial learning rate of 3e-5, decayed to a final learning rate of 1e-9 for the intermediate-task and 1e-10 for the target task. A dropout of 0.25 is introduced between model layers to address overfitting. Due to imbalanced class distributions, class weights are incorporated during training to enhance model generalization on underrepresented classes. Experimental setups adhere to the original papers for baseline models unless otherwise specified, in which case our experimental configurations are adopted.

\subsection{Evaluation metrics}
For consistency with prior works~\cite{ghosh2019stance}\cite{augenstein2016stance}\cite{mohammad2016semeval}, the evaluation of our model employs the macro-average F1-score for the \textit{InFavor} and \textit{Against} classes.

\subsection{Results} \label{res}
%New Table
\begin{table*}[ht]
\caption{Experimental results without sarcasm detection pre-training}
\begin{center}
\begin{small}
%\centering
\begin{tabular}{|l||c|c|c|c|c|c||c|c|c|c|c|c|}
\hline
\multirow{2}{*}{\textbf{Model}}&\multicolumn{6}{|c|}{SemEval}&\multicolumn{6}{|c|}{MPCHI}\\% &\multicolumn{10}{|c|}{MPCHI}\\
\cline{2-13} 
&\textbf{AT}&\textbf{CC}&\textbf{FM}&\textbf{HC}&\textbf{LA}&\textbf{Avg}&\textbf{MMR}&\textbf{SC}&\textbf{EC}&\textbf{VC}&\textbf{HRT}&\textbf{Avg}\\
\hline
Sem-TAN-&0.596 &0.420 &0.495 &0.543 &0.603 &0.531 &0.487 &0.505 &0.564 &0.487 &0.467 &0.502\\
\hline
Sem-CNN&0.641 &0.445 &0.552 &0.625 &0.604 &0.573 &0.524 &0.252 &0.539 &0.524 &0.539 &0.476 \\ 
\hline
Com-BiLSTM&0.567 &0.423 &0.508 &0.533 &0.546 &0.515 &0.527 &0.522 &0.471 &0.474 &0.469 &0.493 \\ 
\hline
ZSSD &0.565 &0.389 &0.546 &0.545 &0.509 &0.511 &- &- &- &- &- &-\\
\hline
Com-BERT&0.704 &0.466 &0.627 &0.620 &0.673 &0.618 &0.701 &0.691 &0.710 &0.617 &0.621 &0.668\\
\hline
ChatGPT &- &- &0.690 &\textbf{0.780} &0.593 &0.687 &- &- &- &- &- &-\\
\hline
Ours-RoBERTa&0.740 &\textbf{0.775} &0.689 &0.683 &0.696 &0.712 &0.692 &0.687 &0.700 &0.701 &0.698&0.695 \\
\hline
Ours-BERT&\textbf{0.767} &0.755 &\textbf{0.697} &0.704 &\textbf{0.702} &\textbf{0.725} &\textbf{0.747} &\textbf{0.722 } &\textbf{0.704} &\textbf{0.702 } &\textbf{0.732} &\textbf{0.721} \\ 
\hline
\end{tabular}
\end{small}
\end{center}
\label{tab:res1}
\end{table*}
%\end{multicols}

%New Table
\begin{table*}[ht]
\caption{Experimental results with sarcasm-detection pre-training}
\begin{center}
\begin{small}
%\centering
\begin{tabular}{|l||c|c|c|c|c|c||c|c|c|c|c|c|}
\hline
\multirow{2}{*}{\textbf{Task}}&\multicolumn{6}{|c|}{SemEval}&\multicolumn{6}{|c|}{MPCHI}\\
\cline{2-13} 
&\textbf{AT} &\textbf{CC} &\textbf{FM} &\textbf{HC} &\textbf{LA} &\textbf{Avg} &\textbf{MMR} &\textbf{SC} &\textbf{EC} &\textbf{VC} &\textbf{HRT} &\textbf{Avg}\\
\hline
SaV2C &0.595  &0.718  &0.596  &0.645  &0.578  &0.626 &0.605 &0.545 &0.545 &0.352 &0.495 &0.508 \\
\hline
SARC &0.697  &0.612  &0.683  &0.557  &0.641  &0.638 &0.605
&0.545 &0.545 &0.352 &0.495 &0.508\\ 
\hline
ST & \textbf{0.769} &\textbf{0.800 }  & \textbf{0.774} &\textbf{0.795}  &\textbf{0.741}  &\textbf{0.775} &\textbf{0.749} &\textbf{0.727} &\textbf{0.704} &\textbf{0.703} &\textbf{0.739} &\textbf{0.724}\\
\hline
\end{tabular}
\end{small}
\end{center}
\label{tab:res2}
\end{table*}

%New table
\begin{table*}[ht]
\caption{Experimental results of an ablation study}
\begin{center}
\begin{small}
%\centering
\begin{tabular}{|l||c|c|c|c|c|c||c|c|c|c|c|c|}
\hline
\multirow{2}{*}{\textbf{Model}}&\multicolumn{6}{|c|}{SemEval}&\multicolumn{6}{|c|}{MPCHI}\\
\cline{2-13} 
&\textbf{AT} &\textbf{CC} &\textbf{FM} &\textbf{HC} &\textbf{LA} &\textbf{Avg} &\textbf{MMR} &\textbf{SC} &\textbf{EC} &\textbf{VC} &\textbf{HRT} &\textbf{Avg}\\
\hline
BERT &0.674  &0.677 &0.678 &0.609 &0.685 &0.665 &0.568 &0.519 &0.441 &0.482 &0.595 &0.521\\
\hline
BERT+Conv+BiLSTM &0.767 &0.755 &0.697 &0.704 &0.702 &0.725 &0.747 &0.722 &\textbf{0.704} &0.702 &0.732 &0.721\\ 
\hline
ST+BERT &0.712 &0.735 &0.698 &0.687 &0.696 &0.706 &0.687 &0.601 &0.540 &0.466 &0.546 &0.568\\
\hline
ST+BERT+Conv&\textbf{0.770} &0.759 &0.689 &0.683 &0.694 &0.719 &0.458 &0.535 &0.479 &0.350 &0.524 &0.469\\
\hline
ST+BERT+BiLSTM &0.747 &0.765 &0.675 &0.657 &0.678 &0.704 &0.640
&0.618 &0.573 &0.528 &0.633 &0.598\\
\hline
ST+BERT+Conv+BiLSTM & 0.769 &\textbf{0.800 }  & \textbf{0.774} &\textbf{0.795}  &\textbf{0.741}  &\textbf{0.775} &\textbf{0.749} &\textbf{0.727} &\textbf{0.704} &\textbf{0.703} &\textbf{0.739} &\textbf{0.724}\\
\hline
\end{tabular}
\end{small}
\end{center}
\label{tab:res3}
\end{table*}

We report averaged results from five experiment runs on each target task. Table~\ref{tab:res1} shows experimental outcomes before sarcasm pre-training in our model. Results for ChatGPT and ZSSD are directly transcribed from their original papers, while the results for other baseline models have been replicated in our experimentation. The table illustrates the commendable performance of our BERT-based model across various targets, with notable superiority in all aspects except for HC and CC, where ChatGPT and our RoBERTa-based model excel, respectively. Consequently, we opt to proceed with our BERT-based model in subsequent experimental results.

Table~\ref{tab:res2} presents experimental results involving sarcasm detection pre-training with our model only. Model performance improves by \textbf{0.050 }and \textbf{0.003} on SemEval and MPCHI, respectively, when pre-trained with ST, surpassing all baseline models in Table~\ref{tab:res1}, but diminishes with Sav2C and SARC.

Table~\ref{tab:res3} presents results of an ablation study using ST only. Different base model components were systematically removed to assess the contribution of each constituent module to the entire model framework. As shown in the table, the model with all components—BERT, Conv, BiLSTM, and sarcasm pre-training—performs the best with average F1-scores of \textbf{0.775} and \textbf{0.724} on SemEval and MPCHI, respectively.

\subsection{Failure Analysis and Discussion} \label{Dis}
Subsequent to obtaining results in Table~\ref{tab:res1}, a failure analysis was conducted on misclassified test samples. Predominantly, misclassifications on SemEval were associated with texts containing sarcastic content, aligning with prior findings~\cite{ghosh2019stance}. This observation substantiated the motivation for considering sarcasm-detection pre-training before fine-tuning on SD.  On the contrary, misclassifications on MPCHI were associated with samples encompassing colossal and generic health-related facts neutral to the respective target under study. Additional observations stemming from the experiments and results across all tasks are outlined below. 

\subsubsection{
Our model outperforms SOTA models even without sarcasm detection}
Specifically, it outperforms ChatGPT and Com-BERT, the best models, on SemEval and MPCHI, by \textbf{0.038} and \textbf{0.053 }on average F1-scores, respectively. While Com-BERT employs only BERT and a dense layer as a classifier, our model incorporates Conv and BiLSTM before the dense layer, contributing to the observed performance improvement. Additionally, it was noted that the inclusion of the BiLSTM module in our model yielded better performance than using pooling layers after the Conv module. This suggests the effectiveness of our model architecture and its ability to capture nuanced representations, leading to proper generalization on SD tasks.

\subsubsection{Sarcasm detection is correlated with SD}
 Consider an illustrative misclassified example: ``\textit{I like girls. They just need to know their place. \#SemST}", a sarcastic comment from the FM target in SemEval. The ground truth for this example is \textit{Against}, but it was predicted as \textit{InFavor} before sarcasm-detection pre-training. Notably, most sarcastic samples in the \textit{Against} class were misclassified as \textit{InFavor} due to their explicit positive content. After incorporating sarcasm knowledge into the model through pre-training, 85\% of misclassified sarcastic samples were predicted correctly.  This observation underscores the relevance of sarcasm-detection pre-training in improving the performance of SD models in our experimentation.

\subsubsection{Not every sarcasm detection  model is a good candidate for intermediate-task transfer learning on SD} 
The inclusion of SARC and SaV2C knowledge in the model pipeline introduced noise and adversely affected model performance on SD compared to incorporating ST knowledge. An analysis of Sav2C and SARC revealed several discrepancies between the intermediate-task datasets and the target tasks. Firstly, the average sentence length in Sav2C and SARC is longer than in SemEval and MPCHI. Secondly, SARC is sourced from different domains than both SemEval and MPCHI, leading to disparities in topic coverage, vocabulary overlap, and the framing of ideas across datasets. Additionally, SARC, being the largest intermediate task, covers a wide range of topics through various subreddits. In contrast, ST, the best-performing intermediate task, shares a similar average sentence length with the target tasks. Moreover, both ST and SemEval are crowd-sourced from Twitter, which likely contributes to the strong performance observed when using ST as an intermediate task on the SemEval dataset. Consequently, the mismatched attributes render certain intermediate tasks less commensurated and less correlated with target tasks, resulting in a negative impact on model performance. Careful consideration and experimentation are essential when selecting a suitable sarcasm model for transfer learning in the context of SD.

\subsubsection{Ablation study regarding sarcasm knowledge}
  The variations in the ablation study results in Table~\ref{tab:res3} help to isolate the effects of each module and determine their individual contributions to the overall improvement in SD performance through sarcasm detection pre-training. 
  %The comparison between these configurations provides insights into the significance of leveraging sarcasm knowledge in the model pipeline.  
  Comparing our best average results in Table~\ref{tab:res1} and Table~\ref{tab:res3}, the infusion of sarcasm knowledge significantly enhances model performance on the SemEval task compared to the MPCHI task. The SemEval task comprises extensive opinionated and sarcastic texts. Conversely, the majority of examples in the MPCHI dataset encompass extensive health-related facts, unrelated to specific targets, aside from occasional sarcasm-related expressions. Consequently, there is a modest increase in performance on MPCHI even when sarcasm detection is utilized. This observation prompts the consideration of exploring variants of BERT or RoBERTa embeddings pre-trained on health-related data specifically for SD on MPCHI as a potential avenue for future work.

\section{Conclusion And Future Work} \label{Con}
 In this study, we introduced a transfer-learning framework that leverages sarcasm detection for SD. RoBERTa and BERT were individually fine-tuned and sequentially concatenated with other deep neural networks, with BERT delivering promising results. The model underwent separate pre-training on three sarcasm-detection tasks before fine-tuning on two target SD tasks. Evaluation against SOTA models demonstrated superior performance, even prior to incorporating sarcasm knowledge. We established the correlation between sarcasm detection and SD, with the infusion of sarcasm knowledge boosting model performance, accurately predicting 85\% of misclassified samples in the SemEval task. Failure analysis revealed SemEval's abundance of opinionated sarcastic samples, underscoring the efficacy of sarcasm pre-training, compared to MPCHI, characterized by generic health-related facts unrelated to specific targets. Additionally, we showed that not every sarcasm-detection intermediate task improved SD due to incongruous linguistic attributes. Finally, an ablation study highlighted that optimal model performance is achieved when utilizing all model constituents.

To the best of our knowledge, this is the  inaugural exploration of sarcasm-detection pre-training applied to the BERT(RoBERTa)+Conv+BiLSTM architecture before fine-tuning for SD. Serving as a foundational reference, our approach establishes a baseline for future researchers in this domain. Future investigations will assess variant BERT or RoBERTa embeddings tailored to health-related text data for the MPCHI task. The research will also concentrate on cross-target SD for both tasks and a more comprehensive examination of other intermediate tasks, including sentiment and emotion knowledge.

\section{LIMITATIONS}
Despite the significant advancements this study brings to NLP applied to social media contexts, several limitations merit consideration. Firstly, the extent of model performance enhancement is contingent upon the attributes of both the intermediary sarcasm detection task and the ultimate SD task. The divergence in linguistic characteristics across datasets utilized for sarcasm detection and SD potentially constrains the broader applicability of the study's outcomes. Secondly, although the integration of BERT or RoBERTa with other deep-learning methodologies represents an innovative approach, the intricate nature of the model architecture may present computational resource challenges and interoperability issues in certain contexts. Lastly, the extensive reliance on fine-tuning techniques and specific datasets raises concerns regarding the model's capacity to generalize effectively across diverse text types or domains not encompassed within the training data corpus.

\section*{ACKNOWLEDGMENT}
This work was supported by the National Science Foundation under Award number OIA-1946391, Data Analytics that are Robust and Trusted (DART). We express our sincere appreciation to our three anonymous reviewers for their valuable insights and constructive feedback. Additionally, we extend our gratitude to all individuals who contributed to this study in various capacities.

%\IEEEtriggeratref{26}

%\newpage

\end{document}